# The Optimal Reward Baseline for Gradient-Based Reinforcement Learning

**Lex Weaver**
Department of Computer Science
Australian National University
ACT AUSTRALIA 0200
*Lex.Weaver@cs.anu.edu.au*

**Nigel Tao**
Department of Computer Science
Australian National University
ACT AUSTRALIA 0200
*Nigel.Tao@cs.anu.edu.au*

## Abstract

There exist a number of reinforcement learning algorithms which learn by climbing the gradient of expected reward. Their long-run convergence has been proved, even in partially observable environments with non-deterministic actions, and without the need for a system model. However, the variance of the gradient estimator has been found to be a significant practical problem. Recent approaches have discounted future rewards, introducing a bias-variance trade-off into the gradient estimate. We incorporate a reward baseline into the learning system, and show that it affects variance without introducing further bias. In particular, as we approach the zero-bias, high-variance parameterization, the optimal (or variance minimizing) constant reward baseline is equal to the long-term average expected reward. Modified policy-gradient algorithms are presented, and a number of experiments demonstrate their improvement over previous work.

## 1 INTRODUCTION

There exist a number of reinforcement learning algorithms which learn by climbing the gradient of expected reward, and are thus categorized as *policy-gradient* methods. The earliest of these was REINFORCE, which solved the immediate reward learning problem, and in delayed reward problems it provided gradient estimates whenever the system entered an identified recurrent state (Williams, 1992). A number of similar algorithms followed, including those in (Glynn, 1986; Cao and Chen, 1997; Cao and Wan, 1998; Fu and Hu, 1994; Jaakkola et al., 1995). The technique of discounting future rewards was introduced in (Kimura et al., 1995; Kimura et al., 1997), and its effect on reducing variance was noted in (Marbach and Tsitsiklis, 1998). The GPOMDP and OLPOMDP algorithms in (Baxter and Bartlett, 1999; Baxter et al., 1999) are the most recent, and remove the reliance on both a system model and a need to identify a specific recurrent state, and operate in partially observable environments with non-deterministic actions (POMDPs).

However, the variance of the gradient estimator remains a significant practical problem for policy-gradient applications, although discounting is an effective technique. Discounting future rewards introduces a bias-variance trade-off: variance in the gradient estimates can be reduced by heavily discounting future rewards, but the estimates will be biased; the bias can be reduced by not discounting so heavily, but the variance will be higher. Our work complements the discounting technique by introducing a *reward baseline*[1] which is designed to reduce variance, especially as we approach the zero-bias, high-variance discount factor.

The use of a reward baseline has been considered a number of times before, but we are not aware of any analysis of its effect on variance in the context of the recent policy-gradient algorithms. (Sutton, 1984) empirically investigated the inclusion of a reinforcement comparison term in several stochastic learning equations, and argued that it should result in faster learning for unbalanced reinforcement tasks. He proposed (and it is widely believed, though without proof) that the average reward would be an intuitive and good value for the comparison term, but he did not provide an analytical reason for it. Williams subsequently demonstrated further algorithms that benefitted from this approach (Williams, 1988), and proved that a baseline did not introduce any bias. Dayan's work (Dayan, 1990) is more closely related to what we present here, in that he explicitly considered the effect of the baseline on variance, although he limits himself to '2-armed bandits', or binary immediate reward problems. His result is summarized in section 2.1.

In the context of the more recent policy-gradient algorithms; (Williams, 1992) focuses on the immediate reward problem, and whilst noting that the different reward baselines did not introduce bias, he found "no basis for choosing among various choices of reinforcement baseline." (Marbach and Tsitsiklis, 1998) employ a differential reward, which is the expectation of the sum of the differences between the rewards received and the average reward (*i.e.* they use an average reward baseline) over the sequence of states between the current and the next occurrence of the identified recurrent state. The use of the average reward $\bar{r}$ as the baseline $b$ has some nice mathematical properties, such as the differential reward of the recurrent state becomes zero, and that the value function $\mathbf{E}[\sum_{t=t'}^{\infty}(R_t - b)]$ is well defined. They note that the choice of an alternative baseline introduces a bias which is linear in the difference $(b - \bar{r})$. However, they fail to notice that the linearity coefficient is a sum of terms of the form $\mathbf{E}[\sum \frac{\nabla P_{ij}}{P_{ij}}]$, which is shown in the next section to be zero. The recent results of Baxter *et. al.* do not consider a reward baseline in their analysis or algorithms (Baxter and Bartlett, 1999; Baxter et al., 1999).

---
[1] Also called *reinforcement comparison* and *reference reward*.

## 2 BACKGROUND

### 2.1 THE IMMEDIATE REWARD PROBLEM

Consider a scenario where a connectionist network maps an input vector $x$ to an output vector $a$, which then determines the reward $r$. The input-output mapping is stochastic, and let $\mu(a|x,\theta)$ denote the probability of the output vector being $a$, given that the input vector is $x$, and some parameter $\theta$ (*e.g.* link weights for a neural network).

The learning task is to find a $\theta$ which maximizes the expected reward $\bar{r} := \mathbf{E}[r|\theta]$, and gradient ascent in $\theta$-space is appropriate, provided that $\nabla \bar{r}$ can be accurately estimated. (Williams, 1992) introduced the REINFORCE algorithm, and showed that, for an arbitrary constant $b$

$$\mathbf{E}[(r - b).\nabla(\log \mu)] = \nabla \bar{r}$$

Note that $(\nabla \log \mu)(a) = \frac{\nabla \mu(a)}{\mu(a)}$, and that

$$\mathbf{E}[b.\frac{\nabla \mu}{\mu}] = b.\sum_a \mu(a).\frac{(\nabla \mu)(a)}{\mu(a)} = b.\sum_a (\nabla \mu)(a)$$

which is zero, since $\mu$ is a probability mass function, so $\sum_a \mu(a) = 1$, and therefore $\sum \nabla \mu = \nabla \sum \mu = \nabla 1 = 0$.

Thus, the choice of $b$, known as the reward baseline, does not bias the gradient estimate $(r - b).\nabla(\log \mu)$. However, it does have an effect on the variance of the estimate. The optimal reward baseline $b^*$ is the one which minimizes this variance. (Dayan, 1990) considered the problem of determining $b^*$ for a binary (2-output) system, and found that

$$b^* = \mu(a_0).\mathbf{E}[r|a_1] + \mu(a_1).\mathbf{E}[r|a_0] \qquad (1)$$

where $a_0, a_1$ are the two possible outputs. His experiments showed that this baseline was better than the more intuitive $\bar{r} = \mathbf{E}[r] = \mu(a_0).\mathbf{E}[r|a_0] + \mu(a_1).\mathbf{E}[r|a_1]$ proposed by (Sutton, 1984).

### 2.2 THE DELAYED REWARD PROBLEM

A more general reinforcement learning problem is that of a Markov Decision Process (MDP), in which an agent's output (or action) affects not only the next reward, but also subsequent rewards. In our notation, there is a state space $\mathcal{X}$, and an action space $\mathcal{A}$. An agent reacts to the system state $X_1, X_2, X_3, \ldots$ with actions $A_1, A_2, A_3, \ldots$ according to its policy function $\mu$ (with parameter $\theta$):

$$\mu(a|x,\theta) := Prob[\ A_t = a \mid X_t = x\ ]$$

There is also a reward function $\rho$, such that the reward process $R_1, R_2, R_3, \ldots$ is given by $R_t = \rho(X_{t+1})$, which reflects the value of choosing action $A_t$. An agent attempts to learn the $\theta$ which maximizes the expected long-term average reward

$$\bar{r} := \lim_{T \to \infty} \mathbf{E}\left[\frac{1}{T}.\sum_{t=1}^{T} R_t\right]$$

We base our work on the GPOMDP algorithm of (Baxter and Bartlett, 1999; Baxter et al., 1999). They present a gradient estimator, which we denote $G_t$, based on the sequence $X_1, A_1, R_1, X_2, A_2, R_2, \ldots, X_t, A_t, R_t$, and a discount factor $\gamma \in [0, 1)$. In essence, the GPOMDP algorithm maintains an eligibility trace vector $Z_t$, and performs the following updates:

$$Z_{t+1} = \gamma.Z_t + \zeta_{t+1} \qquad (2)$$

$$G_{t+1} = G_t + \frac{1}{t+1}.(R_{t+1}.Z_{t+1} - G_t) \qquad (3)$$

where $\zeta_t$ is a random variable given by $\zeta_t := \frac{(\nabla \mu)(A_t|X_t,\theta)}{\mu(A_t|X_t,\theta)}$. We then have the following two identities:

$$Z_t = \sum_{s=1}^{t} \gamma^{t-s}.\zeta_s \qquad (4)$$

$$G_t = \frac{1}{t}.\sum_{s=1}^{t} R_s.Z_s \qquad (5)$$

Baxter and Bartlett show that $\lim_{\gamma \to 1} \lim_{t \to \infty} G_t = \nabla \bar{r}$ w.p. 1, and that the variance of $G_t$ increases with $\gamma$.

## 3 THE OPTIMAL BASELINE

It is straightforward to include a constant baseline $b$ into the GPOMDP algorithm, in that the update rule (3) is simply replaced by $G_{t+1} = G_t + \frac{1}{t+1}.((R_{t+1} - b).Z_{t+1} - G_t)$ and note that, similar to the immediate reward case, the inclusion of a baseline does not affect $\mathbf{E}[G_t]$, since $\mathbf{E}[\zeta_t] = \mathbf{E}[\mathbf{E}[\zeta_t|X_t]] = 0$, as before, so that $\mathbf{E}[Z_t] = 0$.

The remainder of this section confirms Dayan's '2-armed bandit' result and also motivates the following theorem[2]:

**Theorem 1.** *The constant baseline, $b^*$, which minimizes the variance of the gradient estimate, $G_t$, satisfies $\lim_{t\to\infty} \lim_{\gamma\to 1} b^* = \bar{r}$.*

### 3.1 MOTIVATION OF THEOREM 1

With a constant baseline $b$, equation (5) becomes $G_t = \frac{1}{t} \cdot \sum_{s=1}^{t}(R_s - b).Z_s$. Rewriting this with equation (4) yields that $t.G_t$

$$
\begin{aligned}
&= \begin{array}{l}(R_1 - b).\gamma^0.\zeta_1 + \\ (R_2 - b).\gamma^1.\zeta_1 + (R_2 - b).\gamma^0.\zeta_2 + \\ (R_3 - b).\gamma^2.\zeta_1 + (R_3 - b).\gamma^1.\zeta_2 + (R_3 - b).\gamma^0.\zeta_3 + \\ \cdots\end{array} \\
&= \sum_{s=1}^{t} Q_s
\end{aligned}
$$

where $Q_s$ is the sum of column $i$:

$$
\begin{aligned}
Q_s &= \zeta_s . \sum_{i=s}^{t} (R_i - b).\gamma^{i-s} \\
&= \left[\zeta_s . \sum_{i=s}^{t} R_i.\gamma^{i-s}\right] - \left[b.\zeta_s . \frac{1 - \gamma^{t-s+1}}{1 - \gamma}\right]
\end{aligned}
$$

which gives us that

$$\frac{dQ_s}{db} = -\zeta_s . \frac{1 - \gamma^{t-s+1}}{1 - \gamma} \qquad (6)$$

and recall that $\mathbf{E}[\zeta_s] = 0$, so we have that

$$\frac{d\mathbf{E}[Q_s]}{db} = \mathbf{E}\left[\frac{dQ_s}{db}\right] = -\mathbf{E}[\zeta_s] . \frac{1 - \gamma^{t-s+1}}{1 - \gamma} = 0 \qquad (7)$$

We will now show that under the limits $\gamma \to 1$ and $t \to \infty$, the variance of $Q_s$ is minimized at $b^* = \bar{r}$. The proof of theorem 1 (regarding variance minimization of $G_t = \sum_{s=1}^{t} Q_s$) is more complicated, due to covariance terms $Cov(Q_i, Q_j)$, and is not covered here. The full result can be found in (Weaver and Tao, 2001).

#### 3.1.1 Minimizing the Variance in $Q_s$

Now, $Var[\cdot] = \mathbf{E}[\cdot^2] - \mathbf{E}[\cdot]^2$, and so the $b^*$ which minimizes the variance of $Q_s$ satisfies

$$0 = \frac{dVar[Q_s]}{db} = \frac{d}{db}\left[\mathbf{E}[Q_s^2] - \mathbf{E}[Q_s]^2\right] = \frac{d}{db}\mathbf{E}[Q_s^2] - 0$$

---

[2]This theorem is subject to the same regularity conditions as imposed in (Marbach and Tsitsiklis, 1998; Baxter and Bartlett, 1999).

by equation (7). Our optimality criterion is therefore $0 = \frac{d}{db}\mathbf{E}[Q_s^2] = \mathbf{E}\left[\frac{d}{db}(Q_s^2)\right] = \mathbf{E}[2.Q_s . \frac{dQ_s}{db}]$. Substituting equation (6) and canceling out the constant terms gives

$$0 = \mathbf{E}\left[\zeta_s^2 . \left(\sum_{i=s}^{t}(b^* - R_i).\gamma^{i-s}\right)\right] \qquad (8)$$

#### 3.1.2 Binary Immediate Reward Problems

For an immediate reward problem we have $\gamma = 0$ and $t = 1$. In this case, equation (8) simplifies to $0 = \mathbf{E}\left[\zeta_1^2 . (b^* - R_1)\right]$. Now, for a binary system (with exactly two possible actions $a_0$ and $a_1$), we have that $\mu(a_0) + \mu(a_1) = 1$, and that $\zeta(a_1) = \frac{(\nabla\mu)(a_1)}{\mu(a_1)} = \frac{-(\nabla\mu)(a_0)}{1-\mu(a_0)}$. Rewriting $\mathbf{E}[\cdot]$ as $\sum_a \mu(a).\mathbf{E}[\cdot | A_1 = a]$ and some straightforward algebraic calculations reduce the optimality criterion to equation (1), which is Dayan's basic result.

#### 3.1.3 The Optimal Baseline as $\gamma \to 1$

We can rewrite equation (8) with $V_t := R_t - \bar{r}$, and then solve for $b^* - \bar{r}$, taking the limit $t \to \infty$, which gives

$$
\begin{aligned}
b^* - \bar{r} &= \frac{\mathbf{E}\left[\zeta_s^2 . \sum_{i=s}^{\infty} V_i.\gamma^{i-s}\right]}{\mathbf{E}\left[\zeta_s^2 . \sum_{i=s}^{\infty} 1.\gamma^{i-s}\right]} \\
&= \frac{\mathbf{E}\left[\zeta_s^2 . (1 - \gamma) . \sum_{i=s}^{\infty} V_i.\gamma^{i-s}\right]}{\mathbf{E}\left[\zeta_s^2\right]}
\end{aligned}
$$

Let $\pi_i$ be the distribution of $X_i$, and let $\hat{\pi}$ be the steady state distribution, which we assume to be unique. We then have $\mathbf{E}[R_i] = \sum_x \pi_i(x).\rho(x)$, and $\bar{r} = \sum_x \hat{\pi}(x).\rho(x)$, where the sums are over all $x \in \mathcal{X}$. Now, regardless of the initial conditions, the probability distribution of the system state $X_i$ tends towards the steady state distribution as $i \to \infty$. We then have $\lim_{i\to\infty} \pi_i(x) = \hat{\pi}(x)$. Consequently,

$$\lim_{i\to\infty} \mathbf{E}[R_i] - \bar{r} = \lim_{i\to\infty} \sum_x \left((\pi_i(x) - \hat{\pi}(x)).\rho(x)\right) = \sum_x 0.\rho(x)$$

so that $\lim_{i\to\infty} \mathbf{E}[|V_i|] = 0$, and so $\forall \epsilon > 0$. $\exists N$. $i \geq N \Rightarrow \mathbf{E}[|V_i|] = O(\epsilon)$. Since we assume $\zeta$ to be bounded, we also have $\mathbf{E}[\zeta_s^2.|V_i|] = O(\epsilon)$.

We can then bound $b^* - \bar{r}$ by

$$
\begin{aligned}
&\frac{1-\gamma}{\mathbf{E}[\zeta_s^2]} . \left(\sum_{i=0}^{N-1} \mathbf{E}[\zeta_s^2 . |V_{i+s}|].\gamma^i + \sum_{i=N}^{t} \mathbf{E}[\zeta_s^2 . |V_{i+s}|].\gamma^i\right) \\
&= \frac{1-\gamma}{\mathbf{E}[\zeta_s^2]} . \left(\sum_{i=0}^{N-1} \mathbf{E}[\zeta_s^2 . |V_{i+s}|].\gamma^i\right) \\
&+ \frac{1-\gamma}{\mathbf{E}[\zeta_s^2]} . \left(\sum_{i=N}^{t} \mathbf{E}[\zeta_s^2 . O(\epsilon)].\gamma^i\right)
\end{aligned}
$$

Now, we have assumed that $\zeta$ and $V$ are bounded, so the first term is bounded by $(1 - \gamma).O(N)$, which goes to 0

as $\gamma \to 1$. The second term is equal to $O(\epsilon)$, regardless of $\gamma$. Thus, as $\gamma \to 1$, the difference between the optimal baseline and the average reward $(b^* - \bar{r})$, can be bounded by an arbitrarily small number. Thus, in the limit, $b^* = \bar{r}$.

## 4 ALGORITHMS

Theorem 1 tells us that as we increase $\gamma$ (thus reducing bias in our gradient estimate), the optimal variance minimizing constant baseline approaches $\bar{r}$. In general, we may not be able to analytically determine $\bar{r}$ for the POMDPs we are working with. However, we can estimate it from simulation by maintaining an adaptive estimate, $B_t$, of $\bar{r}$, such that $B_t = \frac{1}{t} \sum_{i=1}^{t} R_i$.

The GARB algorithm presented in Figure 1 is a variation on the GPOMDP algorithm in (Baxter and Bartlett, 1999), which has been modified to use this online estimate of the average reward as a baseline. The intuition is that as $t \to \infty$, $Var[B_t] \to 0$. Hence, as $t$ grows large, $B_t$ behaves as a constant, and by definition will approach the value $\bar{r}$.

The OLPOMDP algorithm, also presented in (Baxter and Bartlett, 1999), is an online learning version of GPOMDP that does not explicitly store a gradient estimate, but instead updates the policy parameters $\theta$ directly at every time step. It was proven to converge to a local maximum for $\bar{r}$ in (Bartlett and Baxter, 2000). Applying the modification mentioned above to OLPOMDP yields the OLGARB algorithm, shown in Figure 2.

## 5 EXPERIMENTS

### 5.1 THE 3-STATE SYSTEM

In (Baxter et al., 1999), a simple three-state system is used to demonstrate that the gradient estimates produced by GPOMDP converge with respect to the true gradient. Their experiments also illustrate the bias-variance trade-off that is introduced with the use of a discount factor.[3]

#### 5.1.1 Variance Reduction

We have experimented with the same three-state system to compare the gradient estimates generated by GPOMDP and GARB. The two graphs on the left of Figure 3 show that for a mid-range discount factor of 0.4, the relative error in the estimates produced by both algorithms is comparable. The variance in the GARB estimates appears to be slightly lower, but a more interesting point to observe is that despite Theorem 1 only holding in the limit as $\gamma \to 1$, i.e. for $\gamma \gg 0.4$, GARB has not performed worse than GPOMDP. Note that we do not make a claim for this relationship in general, but merely observe that for some systems GARB is

---

GARB: GPOMDP with an Average Reward Baseline.

Given:
- A POMDP with observations $X_0, X_1, X_2, \ldots$
- A controller with policy function $\mu = Prob[A|X, \theta]$

Algorithm parameters:
- $t$: the number of iterations to run for.
- $\gamma$: the discount factor.
- $\theta'$: a specific policy parameterization.

Variables (all initialized to zero):
- $B$: the baseline (estimated average reward).
- $\vec{Z}$: the eligibility trace vector.
- $\vec{G}$: the gradient estimate vector.

---

1. For $s$ from 1 to $t$ do:
   (a) Receive observation $X_s$.
   (b) Choose action $A_s$ according to $\mu(\cdot|X_s, \theta')$.
   (c) Apply action $A_s$.
   (d) Receive reward $R_s$.
   (e) Update variables according to:
   $B := B + (R_s - B)/s$
   $\vec{Z} := \gamma.\vec{Z} + \frac{\nabla_\theta \mu}{\mu}(A_s|X_s, \theta')$
   $\vec{G} := \vec{G} + ((R_s - B).\vec{Z} - \vec{G})/s$

2. Return $\vec{G}$

Figure 1: The GARB algorithm for estimating the gradient of the long term expected average reward, with repsect to policy parameters $\theta$, for a Partially Observable Markov Decision Process (POMDP). The gradient estimate is for a particular point $\theta'$ in $\theta$-space.

---

[3]Bias is measured by relative error, i.e. $\frac{\|G_t - \nabla \bar{r}\|}{\|\nabla \bar{r}\|}$.

competitive with GPOMDP for $\gamma \ll 1$. Of more significance are the two graphs on the right of the figure. With a discount rate of 0.99, the variance exhibited by GPOMDP has a serious detrimental impact on the gradient estimates. The estimates from GARB suffer a noticeable, but nowhere near as detrimental, increase in variance. This accords well with Theorem 1, in that using the average reward as a baseline should yield reduced variance in comparison to a zero baseline (for discount factors approaching 1).

#### 5.1.2 Demonstrating Theorem 1

For our second experiment we replaced the adaptive estimate of $\bar{r}$ in GARB, with constants[4] from the interval $(0, 1.4\bar{r})$, and for each constant and discount factor pairing, we generated 300 gradient estimates (each after 100 iterations of the system) and calculated their relative errors with respect to the true gradient. After calculating the means and standard deviations of these groups of 300 errors, we could determine which constant reward baseline minimizes the mean error and the variance for a given discount factor. Figure 4 shows the results for four values of $\gamma$. The optimal baseline moves with $\gamma$, from about $0.6\bar{r}$ for $\gamma = 0.4$, to very near $\bar{r}$ for $\gamma = 0.99$. This is an explicit demonstration of Theorem 1, showing that the variance minimizing constant reward baseline approaches $\bar{r}$ as the discount factor approaches 1.

These results strongly suggest that GARB will be useful for POMDPs where large discount factors are necessary, due to smaller discount factors introducing an unacceptable level of bias to the gradient estimates.

### 5.2 PUCKWORLD

In (Baxter et al., 1999) the GPOMDP algorithm is successfully demonstrated on the Puckworld system — a continuous task in which the controller has to navigate a puck over a plane surface to a target point. The puck is controlled by applying a 5 unit force in either the positive or negative $x$ direction, and also in either the positive or negative $y$ direction, for a total of four distinct controls. The controller receives a reward at each decision point (every 0.1 seconds) of $-d$ (the distance between the puck and the target). Periodically, both the puck and the target point are transported to new randomly determined locations. The controller is trained by CONJPOMDP, which uses GPOMDP to generate gradient estimates and GSEARCH to find a local maximum in the gradient direction.

Initial experimentation for (Baxter et al., 1999) found that this system was best trained with a discount factor close to 1, indicating significant bias in gradient estimates for lower valued discount factors. Hence, we would now ex-

---

OLGARB: On-Line GARB.

Given:

- A POMDP with observations $X_0, X_1, X_2, \ldots$.
- A controller with policy function $\mu = Prob[A|X, \theta]$

Algorithm parameters:

- $t$: the number of iterations to run for.
- $\gamma$: the discount factor.
- $\alpha$: the step size factor.

Global variables:

- $\theta'$: a specific policy parameterization.

Local variables (all initialized to zero):

- $B$: the baseline (estimated average reward).
- $\vec{Z}$: the eligibility trace vector.

1. For $s$ from 1 to $t$ do:
   (a) Receive observation $X_s$.
   (b) Choose action $A_s$ according to $\mu(\cdot|X_s, \theta')$.
   (c) Apply action $A_s$.
   (d) Receive reward $R_s$.
   (e) Update variables according to:
   $B := B + (R_s - B)/s$
   $\vec{Z} := \gamma.\vec{Z} + \frac{\nabla_\theta \mu}{\mu}(A_s|X_s, \theta')$
   $\theta' := \theta' + \alpha.(R_s - B).\vec{Z}$

Figure 2: The OLGARB algorithm for updating the specific policy parameters, $\theta'$, in the estimated gradient direction. $\theta'$ is a global variable which is modified during the course of this algorithm.

---

[4] $\bar{r}$ can be analytically determined for this system, but for illustrative purposes we express the baselines as multiples of $\bar{r}$.

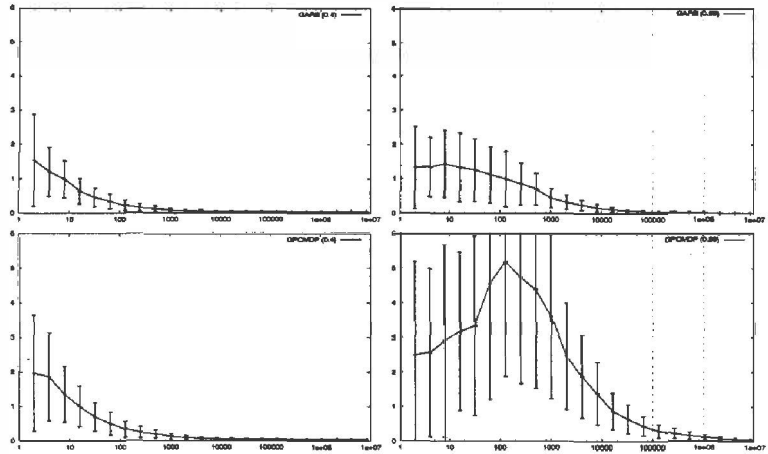

Figure 3: The relative error (bias) in the gradient estimate with respect to the true gradient, graphed against $t$, the number of iterations. Means and standard deviations calculated for 300 independent runs. Note that the improvement of GARB over GPOMDP is more prominent for the higher discount factor $\gamma = 0.99$.

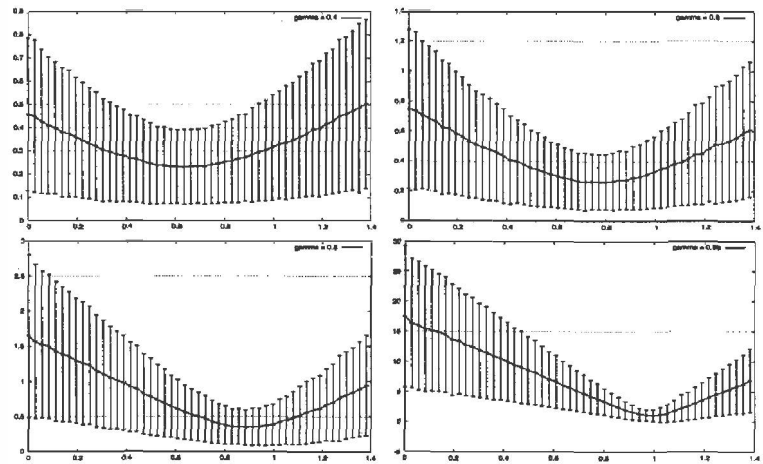

Figure 4: The relative error (bias) in the gradient estimate after 100 iterations with respect to the true gradient, graphed against $b/\bar{r}$, the constant baseline as a proportion of long-term expected average reward. Note that as $\gamma \to 1$, the optimal $b$ approaches $\bar{r}$.

pect the use of GARB to result in more consistent learning. Consequently, we have repeated the previous experiment with CONJPOMDP using each of GPOMDP and GARB to supply gradient estimates, and with a discount factor of 0.95 (as used in (Baxter et al., 1999)). Figures 5 and 6 show the average rewards experienced by controllers trained by CONJPOMDP, with GPOMDP and GARB each supplying the gradient estimates. It is clear from the figures that using GARB yields more consistent learning, both during the initial performance improvement phase, and after the controllers have converged to a good policy. In particular, Figure 7 shows the advantage of GARB in the early stages of training, with there being a statistically significant separation between the means of the GARB and GPOMDP groups.

### 5.3 ACROBOT

In order to demonstrate the OLGARB algorithm, we applied it to the acrobot problem, which is analogous to a gymnast swinging on a high bar. It involves simulating a two-link underactuated robot, with torque applied only at the second joint. Our implementation is based upon the description given in (Sutton and Barto, 1998), with the minor modifications mentioned in (Weaver and Baxter, 1999).

Actions are chosen every 0.1 simulated seconds, though motion is simulated at a much finer granularity. The system is run continuously with reward given after simulating the effect of each action; the reward being simply the height of the acrobot's tip above its lowest possible position. Since both links are 1 metre in length, the reward for each step is between 0 and 4. The controller chooses actions probabilistically according to $\mu(A_s) = \frac{\exp \sum_i \theta_i \cdot \phi_i(X_{s+1}|X_s, A_s)}{\sum_{a \in A} \exp \sum_i \theta_i \cdot \phi_i(a)}$, where $\phi$ is the feature vector $[\phi_1, \phi_2, \phi_3, \phi_4]^T = [|\omega_1|, |\dot{\omega}_1|, |\omega_2|, |\dot{\omega}_2|]^T$.

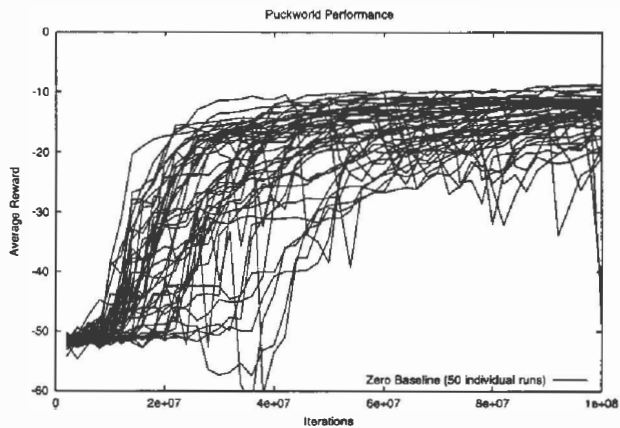

Figure 5: Puckworld controllers trained using GPOMDP.

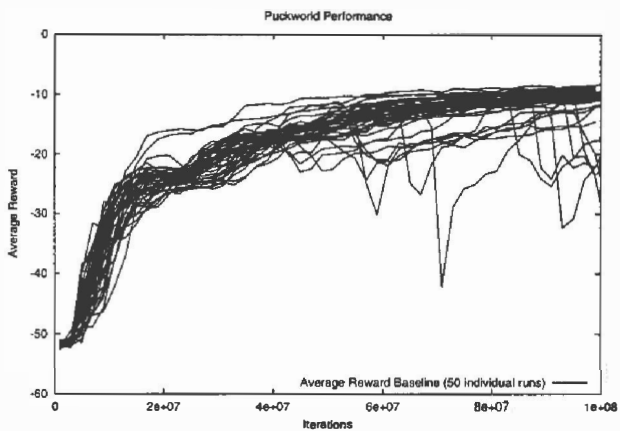

Figure 6: Puckworld controllers trained using GARB.

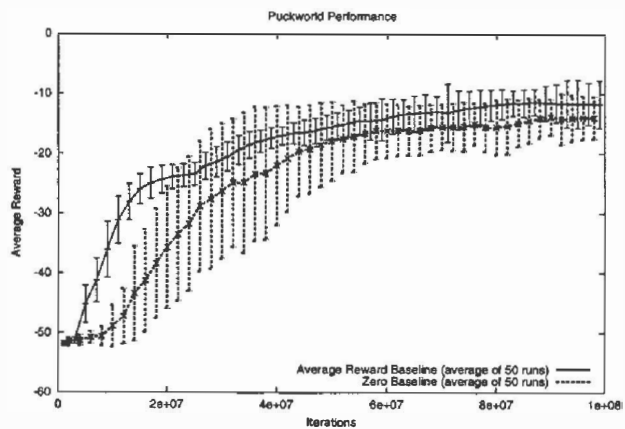

Figure 7: Puckworld: a comparison of CONJPOMDP using GPOMDP and GARB. Error bars are one standard deviation.

Figures 8 and 9 show the average rewards experienced by 100 controllers whose $\theta$ vectors were updated by OLPOMDP and OLGARB respectively. Elements of $\theta$ were initialised randomly to values in $[-0.5, 0.5]$, the step size ($\alpha$) was a 0.01, and the discount factor ($\gamma$) was 0.99. It can readily be seen in the figures that with the exception of a few bad runs, the controllers trained with OLGARB converge well to policies returning an average reward just over 1.2, which is equal to the best hand-coded controller we have developed. In contrast, controllers trained by OLPOMDP converge far less reliably to these good policies. Figure 10 further demonstrates this, showing a comparison of the means and standard deviations of the average rewards experienced by the two groups, with the lower variance of the OLGARB controllers clearly evident.

## 6 CONCLUSION

Several researchers have previously shown that the use of a reward baseline does not bias the gradient estimate, but motivation for using the long-term average expected reward as a baseline has mainly been based on qualitative arguments and empirical success. In section 3.1.2 we show how Dayan's result for binary immediate reward problems (important as a counter example to the intuition mentioned above) is related to, and is thus not inconsistent with, our theorem.

The theorem we have presented provides guidance in choosing a baseline for discount factors approaching 1, by justifying $\bar{r}$ as a variance minimizer. We have demonstrated this using the three-state system, by comparing GARB with a baseline of $\bar{r}$ to GPOMDP (section 5.1), and also by illustrating that the variance minimizing baseline approaches $\bar{r}$ as the discount factor increases (Figure 4).

We have derived variations on existing algorithms which take advantage of theorem 1, and demonstrated the advantage given by the variance reduction these algorithms exhibit. Our demonstrations include calculating and comparing gradient estimates for three-state system, and online policy improvement with the continuous state-space acrobot system. In particular, we have demonstrated, using the Puckworld system, that our result can be of benefit when working with POMDPs that require large discount factors to avoid unacceptable levels of bias.

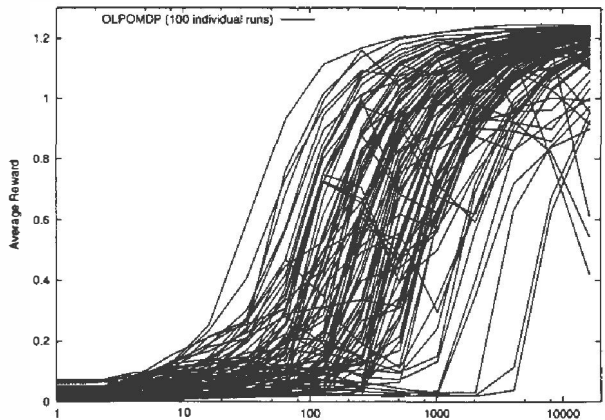

Figure 8: Acrobot controllers trained with OLPOMDP.

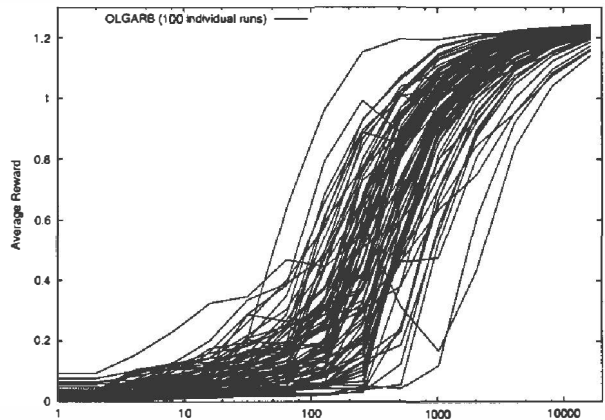

Figure 9: Acrobot controllers trained with OLGARB.

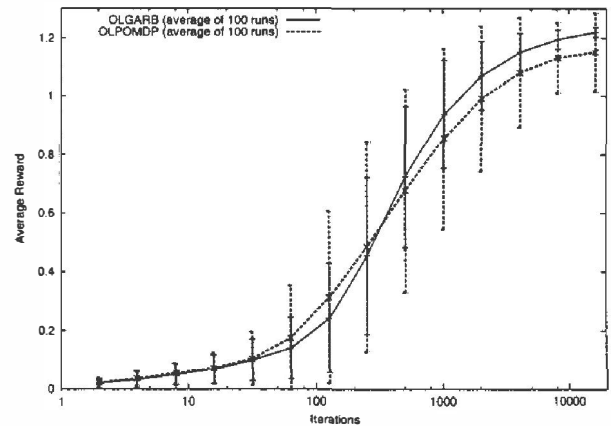

Figure 10: Acrobot: a comparison of OLPOMDP and OLGARB. Error bars are one standard deviation.